%% file: Formatting-Instructions-LaTeX-2022.tex
\def\year{2022}\relax
\documentclass[letterpaper]{article} 
\usepackage{aaai22}  
\usepackage{times}  
\usepackage{helvet}  
\usepackage{courier}  
\usepackage[hyphens]{url}  
\usepackage{graphicx} 
\usepackage{natbib}  
\usepackage{caption} 
\DeclareCaptionStyle{ruled}{labelfont=normalfont,labelsep=colon,strut=off} 
\usepackage{hyperref}       
\usepackage{booktabs}       
\usepackage{amsfonts}       
\usepackage{nicefrac}       
\usepackage{microtype}      
\usepackage{xcolor}         
\usepackage{amsmath}
\usepackage{amssymb}
\usepackage[ruled,linesnumbered,vlined,algo2e]{algorithm2e} 
\usepackage{bm}
\usepackage{color}
\usepackage{epsfig}
\usepackage{diagbox} 
\usepackage{float}
\usepackage{amsthm}
\usepackage{multirow}

\usepackage{algorithm}
\usepackage{algorithmic}

\usepackage{hyperref}
\hypersetup{
    colorlinks=false,
    urlcolor=magenta,
}

\usepackage{newfloat}
\usepackage{listings}
\floatstyle{ruled}
\newfloat{listing}{tb}{lst}{}
\floatname{listing}{Listing}

\title{Boosting Active Learning via Improving Test Performance}
\author {
    Tianyang Wang,\textsuperscript{\rm 1}\thanks{The first two authors contributed equally.}
    Xingjian Li,\textsuperscript{\rm 2 \rm 6}\footnotemark[1] 
    Pengkun Yang,\textsuperscript{\rm 3}
    Guosheng Hu,\textsuperscript{\rm 4} \\
    Xiangrui Zeng,\textsuperscript{\rm 7}
    Siyu Huang,\textsuperscript{\rm 5}
    Cheng-Zhong Xu,\textsuperscript{\rm 6}
    Min Xu\textsuperscript{\rm 7}\thanks{\textbf{corresponding author}}
}
\affiliations {
    \textsuperscript{\rm 1}Austin Peay State University,
    \textsuperscript{\rm 2}Baidu Inc, 
    \textsuperscript{\rm 3}Tsinghua University, \\
    \textsuperscript{\rm 4}Oosto,
    \textsuperscript{\rm 5}Harvard University, 
    \textsuperscript{\rm 6}University of Macau,
    \textsuperscript{\rm 7}Carnegie Mellon University\\
    toseattle@siu.edu, lixingjian@baidu.com, yangpengkun@tsinghua.edu.cn,  huguosheng100@gmail.com, \\ xiangruz@andrew.cmu.edu, huang@seas.harvard.edu, czxu@um.edu.mo, 
    mxu1@cs.cmu.edu
}

\usepackage{bibentry}

\begin{document}

\maketitle

\begin{abstract}
  Central to active learning (AL) is what data should be selected for annotation. Existing works attempt to select highly uncertain or informative data for annotation. Nevertheless, it remains unclear  how selected data impacts the test performance of the task model used in AL. In this work, we explore such an impact by theoretically proving that selecting unlabeled data of higher gradient norm leads to a lower upper-bound of test loss, resulting in a better test performance.  
  However, due to the lack of label information, directly computing gradient norm for unlabeled data is infeasible.
  To address this challenge, we propose two schemes, namely expected-gradnorm and entropy-gradnorm. The former computes the gradient norm by constructing an expected empirical loss while the latter constructs an unsupervised loss with entropy. Furthermore, we integrate the two schemes in a universal AL framework. We evaluate our method on classical image classification and semantic segmentation tasks. 
  To demonstrate its competency in domain applications and its robustness to noise, we also validate our method on a cellular imaging analysis task, namely cryo-Electron Tomography subtomogram classification.
  Results demonstrate that our method achieves superior performance against the state of the art. Our source code is available at \url{https://github.com/xulabs/aitom/blob/master/doc/projects/al_gradnorm.md}.
  
\end{abstract}

\section{Introduction}
\label{intro}
In most scenarios, supervised learning is still the most reliable fashion for training deep neural networks. 
However, data annotation is often arguably expensive. To reduce the cost, active learning (AL) can be used to select a portion of all unlabeled data for annotation and the annotated data is then used to train a task model (e.g. CNN) in a supervised fashion. The goal of AL is to obtain the best test performance of the task model given a specific annotation budget.
Therefore, the question becomes: what data should be selected for annotation to achieve this goal?

According to the fashion for data selection, existing AL methods can be generally categorized into two groups, namely uncertainty-based and diversity-based. The former group aims to select the most uncertain data for annotation (e.g. \cite{caramalau2021sequential, zhang2020state, yoo2019learning, sinha2019variational}),  
whereas the latter group aims to select unlabeled data that can diversify the labeled pool \cite{hasan2015context, yang2015multi, mac2014hierarchical, elhamifar2013convex, bilgic2009link, nguyen2004active}, which can also be achieved by synthesizing new diverse labeled data \cite{mayer2020adversarial, mahapatra2018efficient, zhu2017generative}. 

Although these two main streams of methods define the state-of-the-art baselines, very little work has explored the connection between selected data and the test performance of the task model used in AL. 
Such a connection can be restated as: if an unlabeled sample is selected for annotation and used to train the task model, then how the test performance of the model will be impacted. 
Conversely, this connection can guide us to select unlabeled data which helps improve the test performance. 

This work focuses on exploring the aforementioned connection. 
To do so, we exploit the influence function \cite{koh2017understanding} to analyze how the test performance is impacted if an individual sample is selected for annotation and used in training. 
Our theoretical analysis shows
that \textit{selecting and annotating unlabeled data of higher gradient norm\footnote{Here, the gradient is w.r.t. model parameters rather than input.} will result in a lower upper-bound of test loss}, leading to a better test performance. This finding  naturally leads to a data selection criterion in AL. To our best knowledge, this is the first work that theoretically analyzes how data selection impacts the test performance in AL. Moreover, to leverage our theoretical findings, we propose two independent schemes to compute the gradient norm since it cannot be directly obtained during data selection due to the lack of label information in AL settings. In the first scheme (namely \textbf{expected-gradnorm}), inspired by \cite{cai2017active, cai2013maximizing}, we make use of all possible labels to compute an expected empirical loss that is adopted to compute the gradient norm.
In the second scheme (namely \textbf{entropy-gradnorm}), we use entropy as a loss to compute the gradient norm without resorting to any label information. The rationale of the two schemes is also explained. 
We then propose a universal AL framework to integrate these two schemes. 

\textit{Note that although we aim to boost AL via improving test performance, there is no need to have an access to test data.} Our theoretical analysis is completely based on the assumption that test data is unknown.

To validate our method, we conduct extensive experiments on three different tasks, including image classification, semantic segmentation, and cryo-ET (cryo-Electron Tomography) subtomogram classification \cite{gubins2019shrec,chen2017convolutional} (\textbf{Appendix} A.5), which is an emerging technique for analyzing the machinery of cellular systems. 
Our method outperforms the state-of-the-art AL baselines, 
demonstrating its competency and task-agnostic nature. 


We summarize our main contributions as follows. Firstly, we fundamentally  formulate the test performance in active learning and find that its major impact factor is gradient norm that can effectively guide unlabeled data selection. Secondly, we propose two schemes to compute gradient norm for unlabeled data without resorting to ground-truth label. Thirdly, we demonstrate that the proposed method achieves superior performance on classical computer vision challenges as well as on a domain task in computational biology.

\section{Related Work}
\label{related}

\textbf{Uncertainty-based AL.} Recent advances 
rely on specialized auxiliary models to estimate uncertainty for unlabeled data. For example, \cite{caramalau2021sequential} designs a GCN \cite{kipf2016semi} to select highly uncertain data for annotation. \cite{haussmann2019deep} adopts a Bayesian network to predict uncertainty. \cite{zhang2020state, sinha2019variational, ducoffe2018adversarial} use adversarial learning and multiple auxiliary models (e.g. VAE \cite{kingma2013auto}) to estimate uncertainty. \cite{yoo2019learning} jointly learns the task model with a loss prediction module, which aims to predict the empirical loss for unlabeled data. This loss prediction module also needs customized design and training. While the other methods
do not require additional models, they still suffer from inefficient unlabeled data selection. For instance, \cite{sener2018active, kuo2018cost} need to solve the classical K-center or 0-1 Knapsack problem when selecting unlabeled data, and hence the time complexity is much higher than the methods \cite{yoo2019learning, sinha2019variational} which only need a feed-forward step using trained auxiliary models during data selection. Similarly, several Bayesian methods \cite{gal2017deep,gal2016dropout} suffer from inefficient uncertainty estimation due to the need of thousands of feed-forward operations for each unlabeled sample.

\noindent
\textbf{Diversity-based AL.} 
This type of methods are also known as distribution-based. For example, \cite{yang2015multi} proposes to add a diversity regularizer in the objective function. 
\cite{elhamifar2013convex} uses sample diversity to provide a guidance for solving a convex optimization problem in order to select diverse data for annotation, whereas \cite{nguyen2004active} utilizes clustering to sample diverse data for annotation.
Traditionally, uncertainty and diversity are treated as different criteria for data selection, however, a recent study \cite{loquercio2020general} points out that they are highly correlated. 
In addition, with the aid of generative models (e.g. GAN \cite{goodfellow2014generative} and VAE), new data can be synthesized to diversify the labeled pool \cite{mayer2020adversarial, mahapatra2018efficient, zhu2017generative}. Due to the difficulty and complexity in adversarial training, it needs extra efforts to deploy these methods, especially when data format is changed. For instance, the auxiliary GAN or VAE needs to be redesigned if using 3D data (e.g. cryo-ET subtomogram image, studied in \textbf{Appendix} A.5). 

\noindent
\textbf{Discussion.} For more discussions of the related work, we refer readers to \textbf{Appendix} A.9, in which we discuss how our work differs essentially from \cite{settles2007multiple, roy2001toward}.

\section{Methodology}
\label{methodology}
We firstly explore what data needs to be selected for annotation by theoretically analyzing the connection between selected data and test performance. Then we propose two schemes to compute gradient norm for unlabeled data. Lastly, we present a universal AL framework to leverage our theoretical findings and the two schemes. 

\subsection{What Data to be Selected}
\label{theory}
The main criterion to evaluate an AL method is the test performance of the task model. To our best knowledge, very little work has managed to interpret how selected unlabeled data directly impacts the test performance. This remains challenging since test data is normally unknown at the time of model training and data selection. 
In this work, we aim to draw such a connection \textit{without resorting to test data}. We also aim to use this connection to guide the data selection in AL. Our ultimate goal is to select unlabeled data that makes the task model yield a better test performance. 

According to \cite{koh2017understanding}, we know that given a model $f_{\theta}$, such as a neural network, removing a sample $x$ from its training set will approximately influence the loss at a test sample $x_j$ by
\begin{equation}
\small
\label{eq-3}
  I_{loss}(x, x_{j})=\frac{1}{n} \nabla_\theta L(f_{\theta}(x_{j}))^\top H_{\theta}^{-1} \nabla_\theta L(f_{\theta}(x))\enspace,
\end{equation}
where $n$ denotes the amount of existing training samples, and $f_\theta(\cdot)$ refers to the feed-forward step that yields the logits output of the model $f_\theta$, and
$H_{\theta}=\frac{1}{n} \sum_{i=1}^n \nabla^2_\theta L(f_{\theta}(x_{i}))$ 
is the average Hessian over all training samples. For each training sample, since we want to compute its influence (if removed) on all the samples in a test dataset, we compute the total influence as follows
\begin{equation}
\small
\begin{split}
  \sum_j I_{loss}(x, x_{j})=\frac{1}{n}\sum_j \nabla_\theta L(T^{c+1}(x_{j}))^\top H_{\theta}^{-1} \nabla_\theta L(T^{c+1}(x))\enspace,
\end{split}
\label{eq-4}
\end{equation}
where the task model in AL cycle $c+1$ (i.e. $T^{c+1}$) is used as $f_\theta$ in Eq. \ref{eq-3}, and the index $j$ is over the test set. 
In this section, we assume that data selection occurs in cycle $c$, and the selected data is used in training in cycle $c+1$.

As indicated in \cite{koh2017understanding}, although $I_{loss}(x, x_{j})$ could be negative when a training sample is harmful to an individual $x_j$, $\sum_j I_{loss}(x, x_{j})$ is generally positive. Intuitively, this means removing a training sample will tend to increase the expected test loss. 

In AL settings, assuming the test loss of $T^{c+1}$ is $L^{c+1}_{test}$, if a training sample $x$ is removed from the labeled pool and not involved in 
training $T^{c+1}$,
then the influenced test loss $L'^{c+1}_{test}$ can be computed by  
\begin{equation}
\small
\label{eq-target}
\begin{split}
  L'^{c+1}_{test} &= L^{c+1}_{test} + \sum_j I_{loss}(x, x_{j}) \\
  &= L^{c+1}_{test} + \frac{1}{n} \sum_j \nabla_\theta L(T^{c+1}(x_{j}))^\top H_{\theta}^{-1} \nabla_\theta L(T^{c+1}(x))\enspace.
\end{split}
\end{equation}
Since test data is unknown during data selection in AL, 
directly computing $\sum_j \nabla_\theta L(T^{c+1}(x_{j}))^\top H_{\theta}^{-1} \nabla_\theta L(T^{c+1}(x))$ is intractable. Given $H_{\theta}$ is positive definite by assumption \cite{koh2017understanding}, we can derive Eq. \ref{eq-approx1} by applying the Frobenius norm\footnote{We ignore the norm order in equations to simplify the notations.} on both sides of Eq. \ref{eq-target}, and then we have
\begin{equation}
\small
\label{eq-approx1}
\begin{split}
  L'^{c+1}_{test} &= ||L^{c+1}_{test} + \frac{1}{n}  \sum_j \nabla_\theta L(T^{c+1}(x_{j}))^\top H_{\theta}^{-1} \nabla_\theta L(T^{c+1}(x))|| \\
  & \leq L^{c+1}_{test} + \frac{1}{n}  ||\sum_j \nabla_\theta L(T^{c+1}(x_{j}))^\top H_{\theta}^{-1} \nabla_\theta L(T^{c+1}(x))|| \\
  &= L^{c+1}_{test} + \frac{1}{n}  ||\nabla_\theta L(T^{c+1}(x))^\top \sum_j H_{\theta}^{-1} \nabla_\theta L(T^{c+1}(x_{j}))||  \\
  & \leq L^{c+1}_{test} + \frac{1}{n}  ||\nabla_\theta L(T^{c+1}(x))|| \cdot ||\sum_j H_{\theta}^{-1} \nabla_\theta L(T^{c+1}(x_{j})) ||\enspace.
\end{split}
\end{equation}
For a potential test dataset, $||\sum_j H_{\theta}^{-1} \nabla_\theta L(T^{c+1}(x_{j}))||$ can be regarded as a fixed term. Therefore, the upper-bound of $L'^{c+1}_{test}$ is mainly determined by $||\nabla_\theta L(T^{c+1}(x))||$. However, directly computing $||\nabla_\theta L(T^{c+1}(x))||$ 
is infeasible due to two challenges:
\begin{enumerate}
    \item The model $T^{c+1}$ is not available by the time of data selection in cycle $c$;
    \item $x$ does not have ground-truth label for computing $L$ (addressed in next section). 
\end{enumerate}

Here, we address the first challenge. 
To do so, we involve an approximation that uses $||\nabla_\theta L(T^{c}(x))||$ to bound $||\nabla_\theta L(T^{c+1}(x))||$. This is reasonable since in AL $x$ is used to train $T^{c+1}$ rather than $T^c$, which means $T^c$ has never observed $x$ during its training. 
The approximation is intuitive since the gradient norm w.r.t. the model parameters of a sample (even not employed in training) is likely to decrease as the model better fits the training set. Moreover, although it is not a rigorous guarantee for \emph{any} sample, we show a theoretical support that the gradient norm of the \emph{average} loss is guaranteed to reduce under reasonable assumptions. The detailed proof is given in \textbf{Appendix} A.1. 
With this bound, we can derive Eq. \ref{eq-approx2} from Eq. \ref{eq-approx1}.
\begin{equation}
\small
\label{eq-approx2}
\begin{split}
  L'^{c+1}_{test} & \leq L^{c+1}_{test} + \frac{1}{n} ||\nabla_\theta L(T^{c+1}(x))|| \cdot ||\sum_j H_{\theta}^{-1}\nabla_\theta L(T^{c+1}(x_{j}))|| \\
  & \lesssim L^{c+1}_{test} + \frac{1}{n}||\nabla_\theta L(T^{c}(x))|| \cdot ||\sum_j H_{\theta}^{-1}\nabla_\theta L(T^{c+1}(x_{j}))||\enspace.
\end{split}
\end{equation}
The Eq. \ref{eq-approx2} indicates that in cycle $c+1$, removing a training sample $x$ of higher $||\nabla_\theta L(T^c(x))||$ will result in a higher upper-bound of $L'^{c+1}_{test}$. Therefore, such $x$ should be preserved in the labeled training pool of cycle $c+1$. 
Conversely, from the perspective of cycle $c$, 
during data selection, more unlabeled samples of higher $||\nabla_\theta L(T^c(x))||$ should be selected for annotation and added to the labeled pool for training $T^{c+1}$.
Such a data selection scheme helps $T^{c+1}$ maintain a lower upper-bound of $L'^{c+1}_{test}$. Therefore, we conclude that \textit{unlabeled data of higher gradient norm should be selected for annotation in AL.} 
To better introduce our motivation of leveraging gradient norm in AL, we provide an intuitive explanation in \textbf{Appendix} A.3. 



\subsection{Computing Gradient Norm}
\label{estimation}
The Eq. \ref{eq-approx2} can be used to guide the unlabeled data selection in AL. Specifically, we select unlabeled data $x$ that leads to a higher $||\nabla_\theta L(T^c(x))||$, in order to lower the upper-bound of the test loss, as analyzed in last section. 
However, computing $||\nabla_\theta L(T^c(x))||$ remains challenging since computing the empirical loss $L$ is not viable due to the lack of the label information of $x$. To address this challenge, we propose two schemes to compute $||\nabla_\theta L(T^c(x))||$, namely \textbf{\textit{expected-gradnorm}} and \textbf{\textit{entropy-gradnorm}}, respectively. 
We also analyze the rationale of the schemes in this section.  

\subsubsection{Expected-Gradnorm Scheme.}
\label{expected}
In order to compute $||\nabla_\theta L(T^c(x))||$, we need to compute the loss $L$ first. Then, for neural nets we can back-propagate the loss to obtain the gradient w.r.t the model parameters $\theta$. Since $L$ cannot be computed directly for unlabeled data, inspired by \cite{cai2017active, cai2013maximizing}, we propose to use an expected empirical loss to approximate the real empirical loss. 
Suppose there are $N$ classes in a given unlabeled pool, we use $y_i$ to denote the label of the $i^{th}$ class. Note that $y_i$ is a label candidate of $x$ rather than the ground-truth of $x$.
For each sample $x$, its expected loss can be computed by \begin{equation}
\small
\label{eq:ExpectedGN}
\begin{split}
  L_{exp}(T^c(x))=\sum_{i=1}^N P(y_i|x)L_i(T^c(x), y_i)\enspace,
\end{split}
\end{equation}  
where $P(y_i|x)$ is the posterior obtained using softmax over $T^c(x)$ and $L_i$ is the empirical loss when the $i^{th}$ label candidate is \textit{assumed} to be the ground-truth label of $x$. Note that this scheme does not require the \textit{real} ground-truth of $x$. 

This scheme can be easily used for classification problems because the posterior $P$ is a single vector for each individual data sample. However, for other problems, such as semantic segmentation (i.e. pixel-level classification), this scheme is not an ideal solution. 
It is because in this scheme we need to consider all possible labels for all individual pixels, resulting in a huge number of possibilities (see \textbf{Appendix} A.4 for details), which is intractable in practice. To address this challenge, we propose another scheme in next section to compute gradient norm for unlabeled data.

\subsubsection{Entropy-Gradnorm Scheme.}
\label{entropygrad}

In this scheme, we use the output entropy to compute gradient norm. Specifically, \textit{we use the differentiable entropy of the softmax output of the network as a loss function.} Since computing entropy does not require label, this scheme is more suitable for complex tasks in which assuming label is infeasible, such as the semantic segmentation task. Following the definition of $P(y_i|x)$ and $N$ in Eq.~\ref{eq:ExpectedGN}, the entropy loss for each sample $x$ is defined as
\begin{equation}
\small
\label{eq:EntropyGN}
\begin{split}
  L_{ent}(T^c(x))=-\sum_{i=1}^N P(y_i|x) log P(y_i|x)\enspace.
\end{split}
\end{equation}  
Without an access to 
any label information, 
it is reasonable to choose entropy as a surrogate of the real loss due to two reasons. 
\begin{enumerate}
    \item According to Eq.~\ref{eq-approx2}, adding a training sample of higher gradient norm (if computed by the entropy loss in Eq.~\ref{eq:EntropyGN}) is expected to 
    reduce the entropy of a test sample. In  classification problems, reducing the entropy is usually 
    beneficial to reducing typical supervised losses such as the cross-entropy. For example, in semi-supervised learning ~\cite{berthelot2019mixmatch, grandvalet2005semi}, entropy is adopted as an effective regularizer that can be minimized, naturally leading to the use of unlabeled samples in training.
    \item When the task model is well trained (a good model, e.g. in later AL cycles), entropy well approximates cross-entropy \cite{srinivas2018knowledge}. 
\end{enumerate}
 
To use this scheme, one only needs to compute entropy for output posterior and back-propagate this entropy through the task model to obtain the gradient. 
Despite its simplicity, this scheme can be used for both classification and semantic segmentation, leading to its task-agnostic nature. \textit{Note that the model parameters will not be updated during data selection, no matter which scheme is used to compute gradient.}

\subsubsection{Discussion of Approximations.}
Note that, \textit{both the expected empirical loss defined in Eq.~\ref{eq:ExpectedGN} and the entropy loss in Eq.~\ref{eq:EntropyGN} can be used in Eq.~\ref{eq-approx2} as the loss function $L$}, since the derivation does not require a particular form of the non-negative $L$. They also satisfy the requirements for the loss function employed in the influence function~\cite{koh2017understanding} and for deriving the reduction of average gradient norm (\textbf{Appendix} A.1). Furthermore, we provide quantitative evaluations to validate that more than 90\% of the selected samples indeed have reduced gradient norm 
when the loss functions defined in Eq.~\ref{eq:ExpectedGN} and Eq.~\ref{eq:EntropyGN} are used to compute gradient. This further demonstrates the reliability of the approximation in Eq.~\ref{eq-approx2}. We show the detailed evaluations in \textbf{Appendix} A.2.

\subsection{Proposed Active Learning Framework}
\label{framework}
To leverage our theoretical findings, we develop a universal AL framework to integrate the two schemes introduced in last section. 
Specifically, we train the task model with randomly selected initial annotated data and then use one of the schemes to select unlabeled data for annotation. The newly annotated data is added to the labeled pool and we re-train the task model with the updated labeled pool. Our framework makes minimal changes to the classical AL pipeline, making it easy to be implemented for different scenarios. We summarize our framework in \textbf{Algorithm 1}. 

\begin{algorithm}[htbp]
\small
\SetKwData{P_u}{P_u}\SetKwData{P_v}{P_v}\SetKwData{P_i}{P_i}\SetKwData{P_j}{P_j}
\SetKwData{T1}{t1}\SetKwData{T2}{t2} \SetKwData{E}{E}
\SetKwData{Left}{left}\SetKwData{This}{this}\SetKwData{Up}{up}
\textbf{Input:} 
\\
$\mathcal{T}$: task model;
$\mathcal{U}$: unlabeled pool; $\mathcal{L}$: labeled pool; 
$\mathcal{Y}$: test dataset; 
$\mathcal{C}$: number of AL cycles\;
$\mathcal{E}$: number of epochs within each cycle;
$L$: empirical loss;
$K$: annotation budget in each cycle; 
$x$: each unlabeled sample;
\\\\
\textbf{Output:} 
\\
the task model $\mathcal{T}$\;
\\\\
\Begin{
initialize $\mathcal{T}$; \\
\For{$i\leftarrow 1$ \KwTo $\mathcal{C}$}{
\For{$j\leftarrow 1$ \KwTo $\mathcal{E}$}{train $\mathcal{T}$ with $L$ on $\mathcal{L}$; \\
}  
\uIf{expected-gradnorm}{compute the expected loss $L_{exp}$ for $x$ according to Eq.~\ref{eq:ExpectedGN}; \\
select $K$ samples of the highest $||\nabla_\theta L_{exp}(T^c(x))||$ for annotation;} 
\uIf{entropy-gradnorm}{compute the entropy loss $L_{ent}$ for $x$ according to Eq.~\ref{eq:EntropyGN}; \\
select $K$ samples of the highest $||\nabla_\theta L_{ent}(T^c(x))||$ for annotation;} 
update $\mathcal{L}$ and $\mathcal{U}$, respectively;
}  
return $\mathcal{T}$\;
}
\caption{Proposed Active Learning Framework.}
\label{algo}
\end{algorithm}

\section{Experiments}
\label{empirical}
To evaluate the proposed method, we conduct extensive experiments with AL settings.
For fair comparison, we reproduce the baseline methods following their suggested settings and practices, such as the way of pre-processing input. 
In all the experiments except ImageNet, we run the AL methods for 7 cycles, corresponding to 7 different annotation budgets (i.e. from 10\% to 40\% with an incremental size of 5\%). For the very first cycle, we randomly select 10\% of the data from the unlabeled pool and use the selected data as the initial training data for all the compared methods. Then for each subsequent cycle, we select unlabeled data using a specific AL method, and re-train the task model with the updated labeled pool. Note that the term \textit{Cycle} is different from \textit{Epoch}, since it only corresponds to the annotation budget in AL. All the reported results are averaged over 3 runs. For ImageNet, we conduct the experiments with 5 AL cycles (i.e. annotation budget varies from 10\% to 30\%) and the results are averaged over 2 runs, which are acceptable for very large-scale datasets.


We compare our method with the state-of-the-art AL baselines, including random selection, core-GCN \cite{caramalau2021sequential}, sraal \cite{zhang2020state}, vaal \cite{sinha2019variational}, ll4al \cite{yoo2019learning}, core-set \cite{sener2018active}, QBC \cite{kuo2018cost}, and mc-dropout \cite{gal2016dropout}.  
We denote our method with \textbf{exp-gn} (expected-gradnorm) and \textbf{ent-gn} (entropy-gradnorm), respectively. 
\textit{We also refer readers to the \textbf{Appendix} for the training details (A.8), more experimental analysis (i.e. the biomedical cryo-ET experiment (A.5)), the ablation study (A.6), and the time efficiency of the different methods (A.7).}

\begin{figure*}[t]
    \centering
    \includegraphics[width=0.43\linewidth]{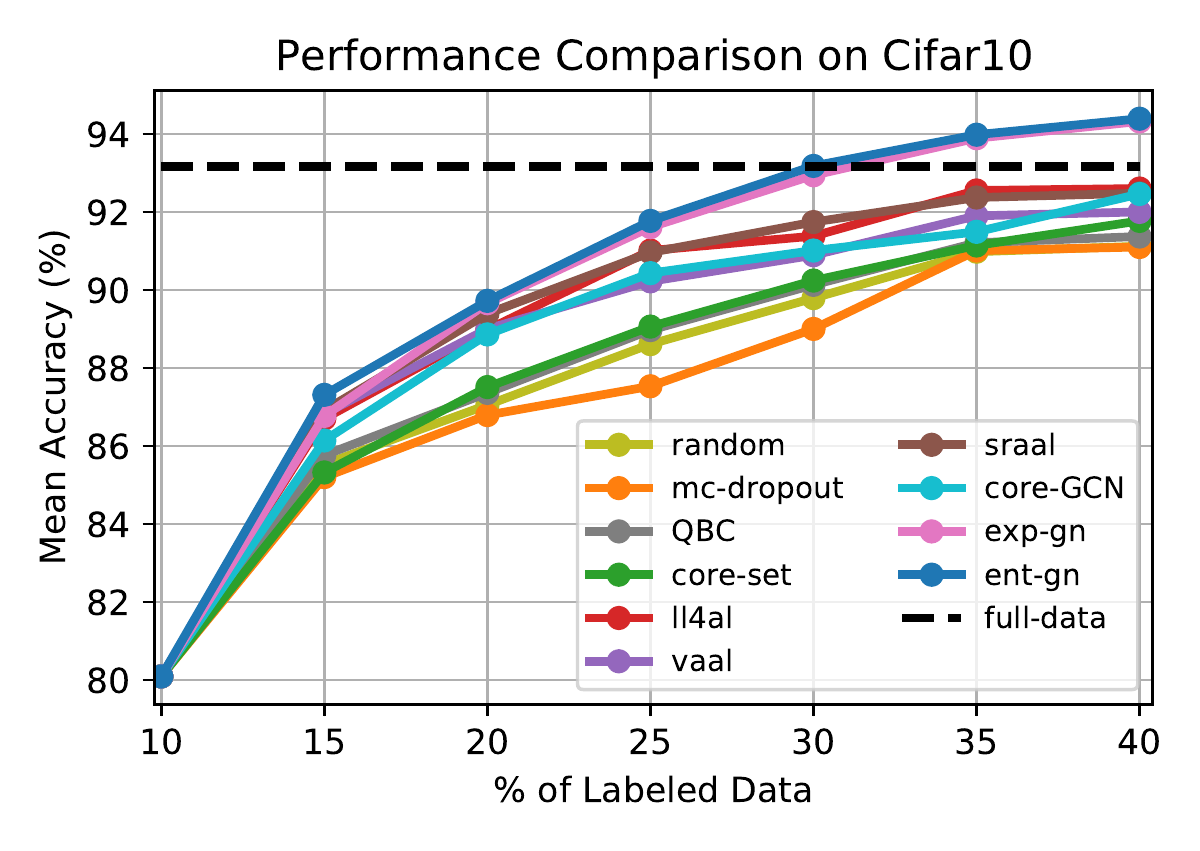}
    \includegraphics[width=0.43\linewidth]{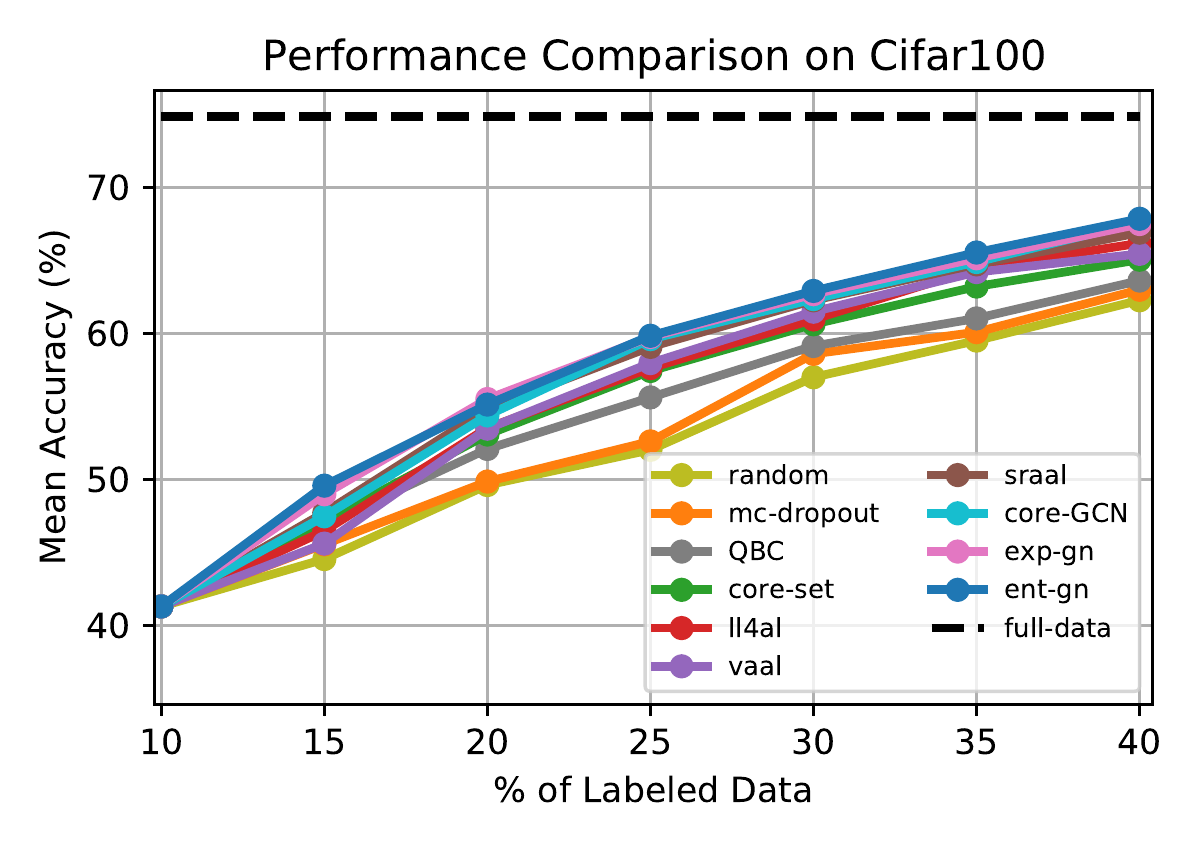}
    \includegraphics[width=0.43\linewidth]{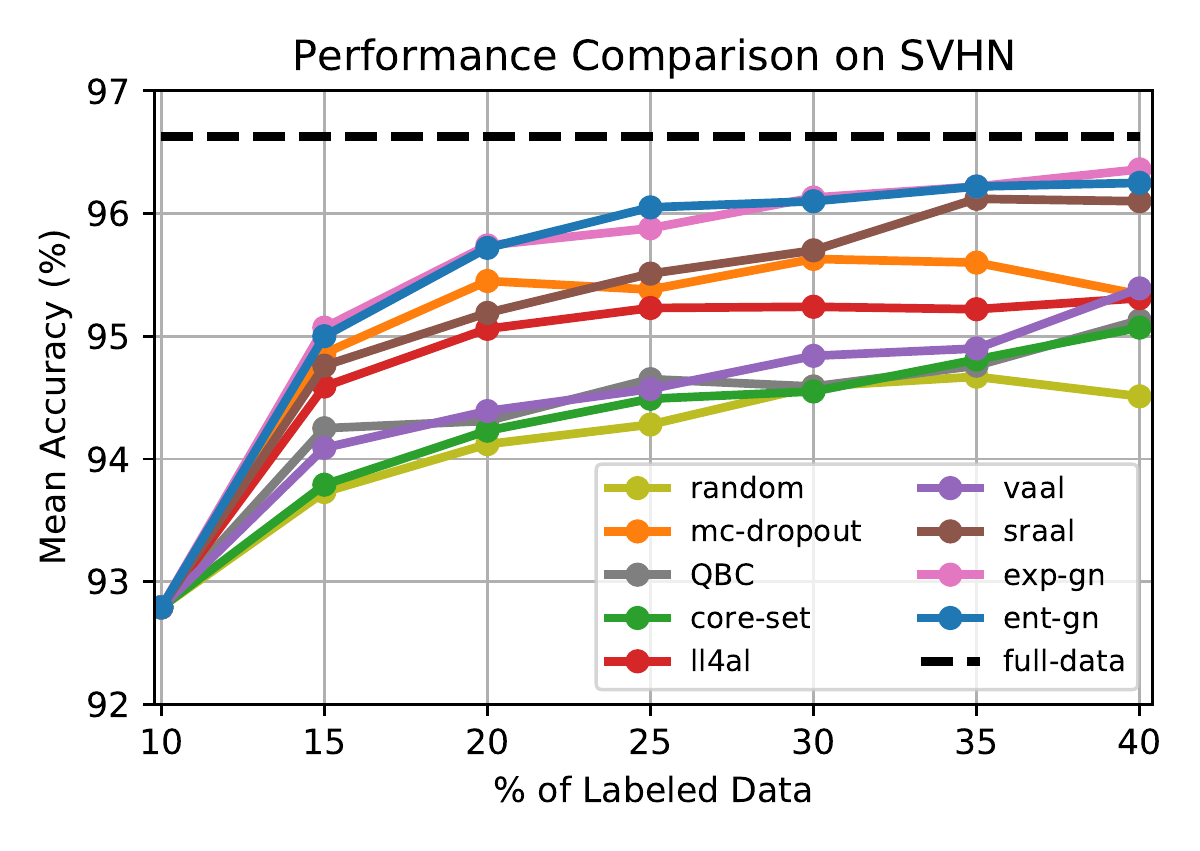}
    \includegraphics[width=0.43\linewidth]{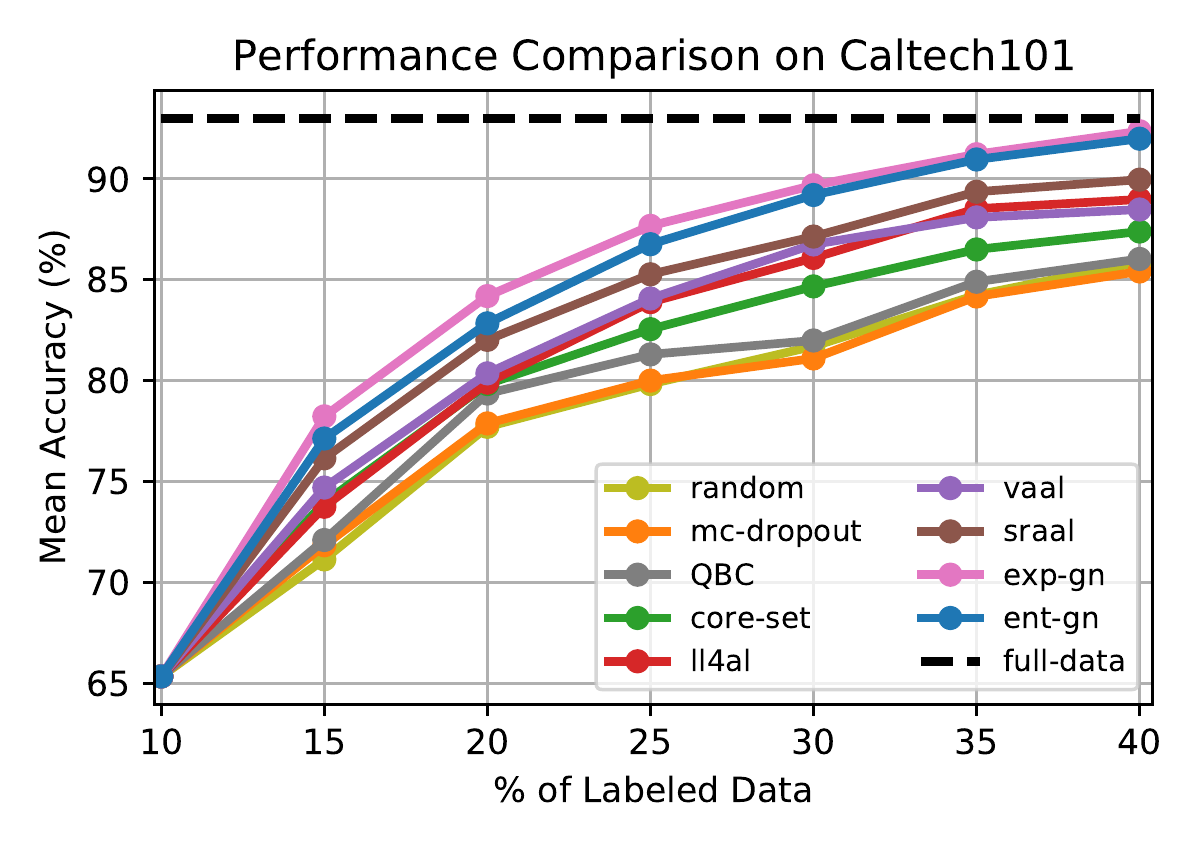}
    
    \caption{Classification performance of the AL methods on Cifar10 (\textbf{top left}), Cifar100 (\textbf{top right}), SVHN (\textbf{bottom left}), and Caltech101 (\textbf{bottom right}), respectively. Dashed line refers to training the model with all the training data. 
    }

    \label{fig:classification1}
\end{figure*}

\subsection{Image Classification}
\label{experiment-classification}

\noindent
\textbf{Datasets.} We exploit five widely used image classification datasets in our experiments, namely Cifar10 \cite{krizhevsky2009learning}, 
Cifar100 \cite{krizhevsky2009learning}, 
SVHN \cite{svhn}, 
Caltech101 \cite{caltech}, 
and ImageNet \cite{deng2009imagenet}. 
Both Cifar datasets include 50000 training and 10000 test samples, distributed across 10 and 100 classes, respectively. SVHN also consists of 10 classes, while it includes more samples than the two Cifar datasets, namely 73257 for training and 26032 for testing. For fair comparisons with the other methods, we do not use the additional training data in SVHN. Caltech101 has 101 classes of bigger images (e.g. $300\times200$ pixels), which are non-uniformly distributed across the classes. Some class has up to 800 samples, while some other only includes 40 samples. 
ImageNet is a large-scale dataset including around 1.28 million training samples across 1000 classes. We follow the common practice to report model performance on the validation set that consists of 50000 samples.

\noindent
\textbf{Model selection.}
\label{modelselect}
Following the practice in \cite{zhang2020state, yoo2019learning}, we use ResNet-18 \cite{he2016deep} as the task model for all the experiments except ImageNet. Specifically, for the two Cifar and SVHN datasets, we exploit a customized version\footnote{https://github.com/kuangliu/pytorch-cifar} of ResNet18, due to the compatibility of input dimensions. 
For Caltech101, we exploit the original ResNet-18. To perform fair comparisons, we use the same task model for all the compared methods. For instance, \cite{sinha2019variational} originally used a VGG Net as the task model, and we replace it with ResNet-18 to maintain consistency. In addition, to validate that our method is \textit{architecture-independent}, we use VGG-16 \cite{simonyan2014very} as the task model for all the methods in the ImageNet task.

\begin{figure*}[t]
    \centering
    \includegraphics[width=0.45\linewidth]{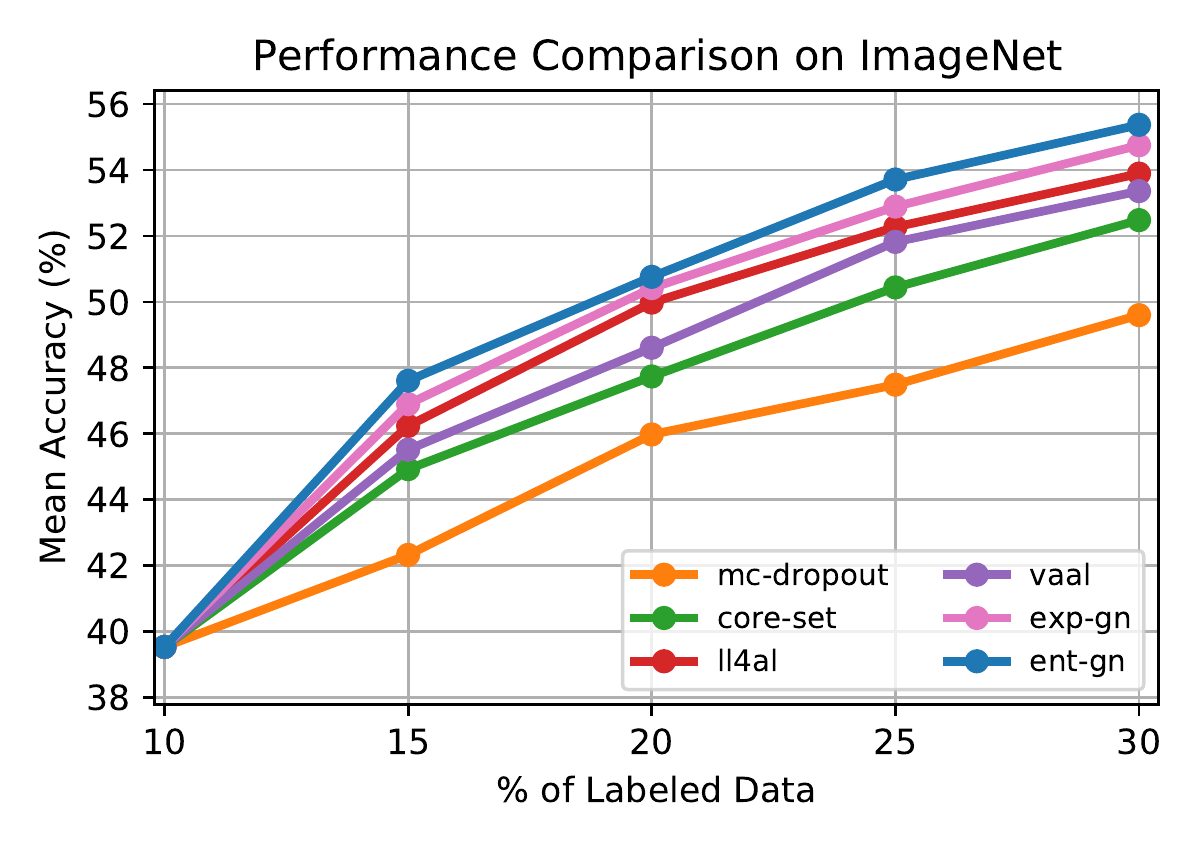}
    \includegraphics[width=0.45\linewidth]{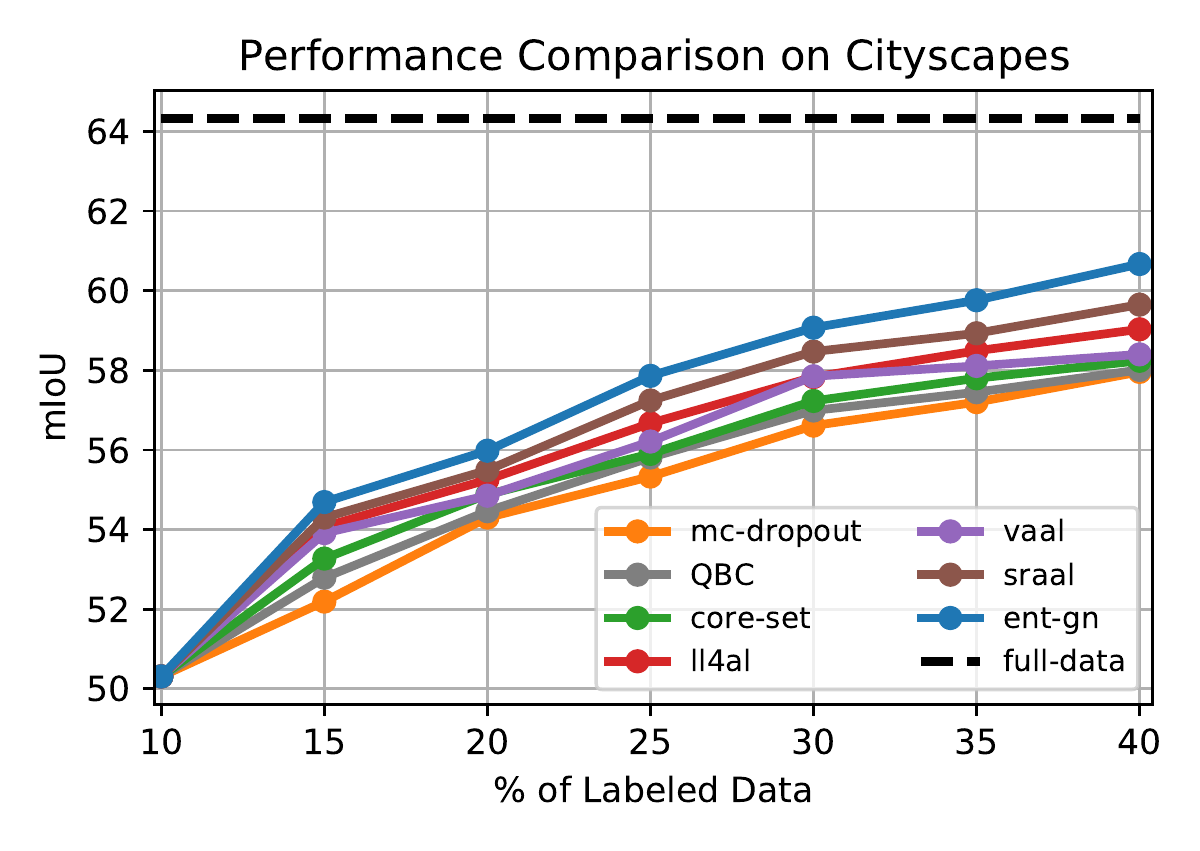}
    
    \caption{Comparison of the AL methods on ImageNet classification (\textbf{left}) and Cityscapes semantic segmentation (\textbf{right}).}
    \label{fig:class-seg}
\end{figure*}

\begin{table*}[th]
\centering
\small
\begin{tabular}{|l|ccccccc|}
\hline
\textbf{budget (\%)} & 10 & 15 & 20 & 25 & 30 & 35 & 40 \\
\hline

\textbf{core-set \cite{sener2018active}} & 217 & 230 & 284 & 339 & 329 & 289 & 255  \\

\textbf{vaal \cite{sinha2019variational}} & 268 & 259 & 215 & 231 & 256 & 240 & 256  \\

\textbf{ll4al \cite{yoo2019learning}} & 391 & 893 & 1084 & 1710 & 1698 & 2035 & 1817  \\

\textbf{sraal \cite{zhang2020state}} & 506 & 819 & 1120 & 1765 & 1842 & 2133 & 2196  \\

\textbf{expected-gradnorm} & \textbf{703} & \textbf{980} & \textbf{1216} & \textbf{1942} & \textbf{2318} & \textbf{2427} & \textbf{2448}  \\

\textbf{entropy-gradnorm} & \textbf{677} & \textbf{932} & \textbf{1141} & \textbf{1993} & \textbf{2317} & \textbf{2429} & \textbf{2443}  \\

\hline
\end{tabular}
\caption{Comparison of the AL methods via the number of the selected samples (out of $K=2500$) of higher gradient norm. The Cifar10 dataset is used.}
\label{table:influence}
\end{table*}

    



\noindent 
\textbf{Results and Analysis.} 
As shown in Fig. \ref{fig:classification1} and Fig. \ref{fig:class-seg} (left), our method outperforms the baselines on all the datasets. Besides, we have the following observations that further demonstrate the advantages of our method. 
Firstly, for each annotation budget, our method yields a higher accuracy than the others. This is a desired property for AL methods since the annotation budget may vary in real-world scenarios. 
Secondly, our method needs fewer labeled samples to achieve a better performance. For example, on Cifar10, our method (ent-gn) yields an accuracy of 91.77\% with 12.5K labeled samples, whereas sraal 
needs 2.5K more and vaal 
needs 5K more samples to achieve the similar performance.
Thirdly, the superior performance on SVHN and Caltech101 demonstrates that our method can well handle imbalanced datasets. 
Fourthly, the superior performance on ImageNet demonstrates that our method is effective for very large-scale datasets.
Lastly, we observe that on Cifar10 our method (ent-gn) achieves an accuracy of 94.39\%, whereas training the same task model from scratch with the full dataset only yields 93.16\%. This interesting finding is aligned with the observation in \cite{koh2017understanding}, which suggests that some training data is harmful to neural network learning.  

\subsection{Semantic Segmentation}
This experiment demonstrates the task-agnostic nature of the proposed method. 

\noindent
\textbf{Dataset.}
We evaluate the AL methods on the Cityscapes \cite{cityscapes} dataset, which is large-scale and consists of video sequences of street scenes from 50 cities. For fair comparison, we only use the standard training and validation data. Following the common practice in \cite{sinha2019variational, yu2017dilated}, we crop the images to a dimension of $688\times688$, and set the number of categories to 19. 

\noindent
\textbf{Model selection.}
\label{modelselect-2}
Following the widely used settings, we adopt the dilated residual network (DRN-D-22) as the task model. 
We refer readers to \cite{yu2017dilated} for the architecture details of this network.


\noindent
\textbf{Results and Analysis.}
We employ the mIoU (mean Intersection-over-Union) to measure the performance of each method. As can be seen in Fig. \ref{fig:class-seg} (right), our method yields consistently better results than the others for all the annotation budgets. In fact, semantic segmentation is a pixel-level classification task, which is more complex than image-level classification. Moreover, the Cityscapes dataset is highly imbalanced. The superior performance demonstrates that our method is robust to imbalanced datasets and suitable for complex tasks. In addition, since our method does not rely on advanced learning fashions (e.g. adversarial learning), there is no need to customize the training scheme with task change, making our method task-agnostic.

\subsection{Our Method Tends to Select More Data of Higher Gradient Norm}
Here, we investigate whether the proposed method tends to select more unlabeled samples of higher gradient norm than the AL baselines. Specifically, for unlabeled pool \{$R_U$\}, we assume that the labels $Y_U$ for all $R_U$ are \textit{temporarily} available and then compute $||\nabla_\theta L(T^{c}(R_U))||$ for each $R_U$. We sort \{$R_U$\} by $||\nabla_\theta L(T^{c}(R_U))||$ in ascending order to form a new set, namely \{$S_U$\}. Then we use each AL method to select $K$ samples \{$X_K$\} from the unsorted unlabeled pool \{$R_U$\}. Afterwards, we compute $len(Set)$, where $Set$ = intersect1d(\{$S_U$\}[-$K$:], \{$X_K$\}) (in Numpy notation) and $len(\cdot)$ yields the length of a set.
As shown in Table \ref{table:influence}, for each annotation budget, our method tends to select more unlabeled samples of higher $||\nabla_\theta L(T^{c}(x))||$ than the others. According to the analysis in \textbf{Methodology}, 
this observation indicates that the task model trained on our selected data can yield better performance on test data, which is aligned with the goal of AL.

\begin{figure*}[th]
    \centering
    \includegraphics[width=0.4\linewidth]{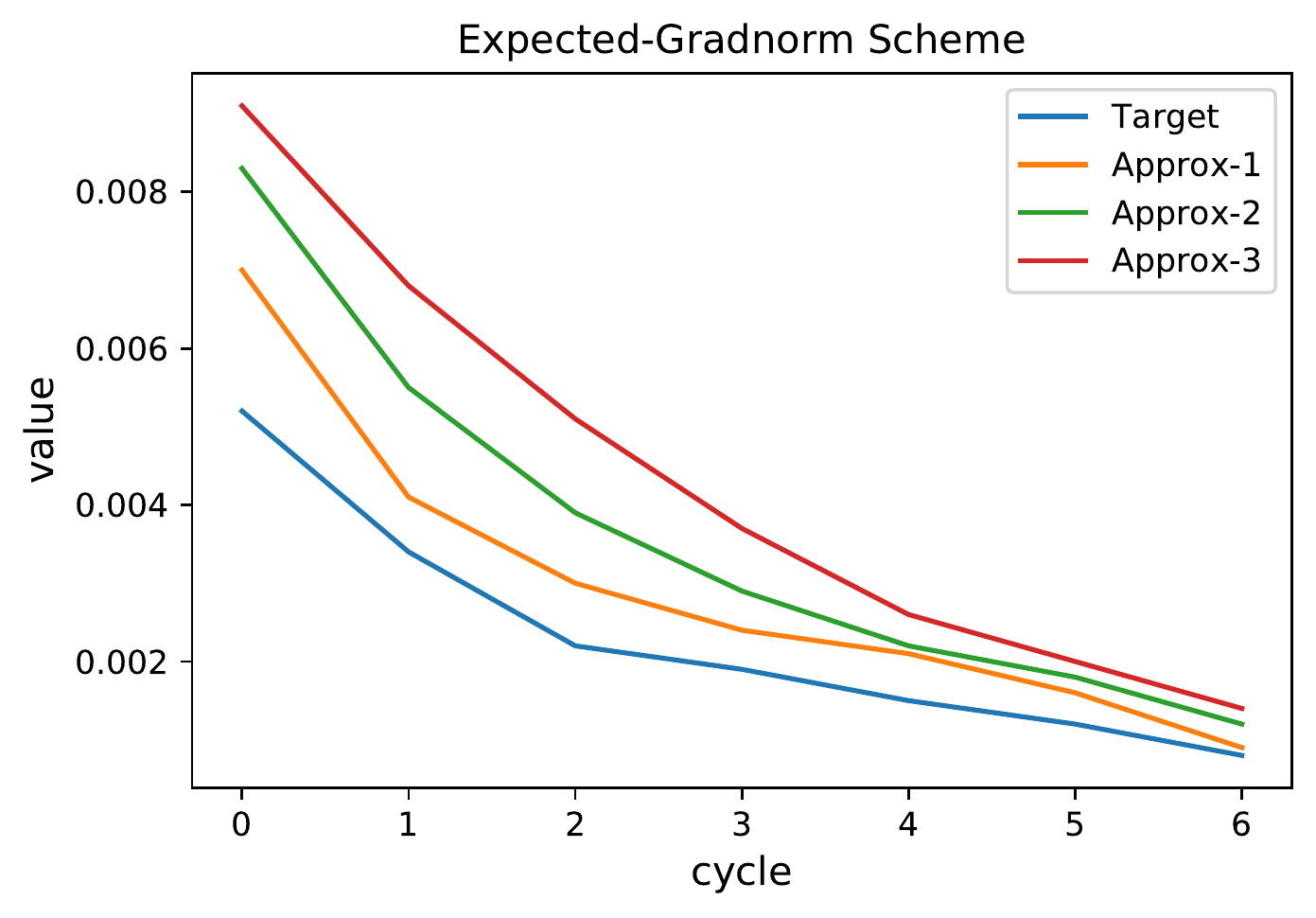}
    \includegraphics[width=0.4\linewidth]{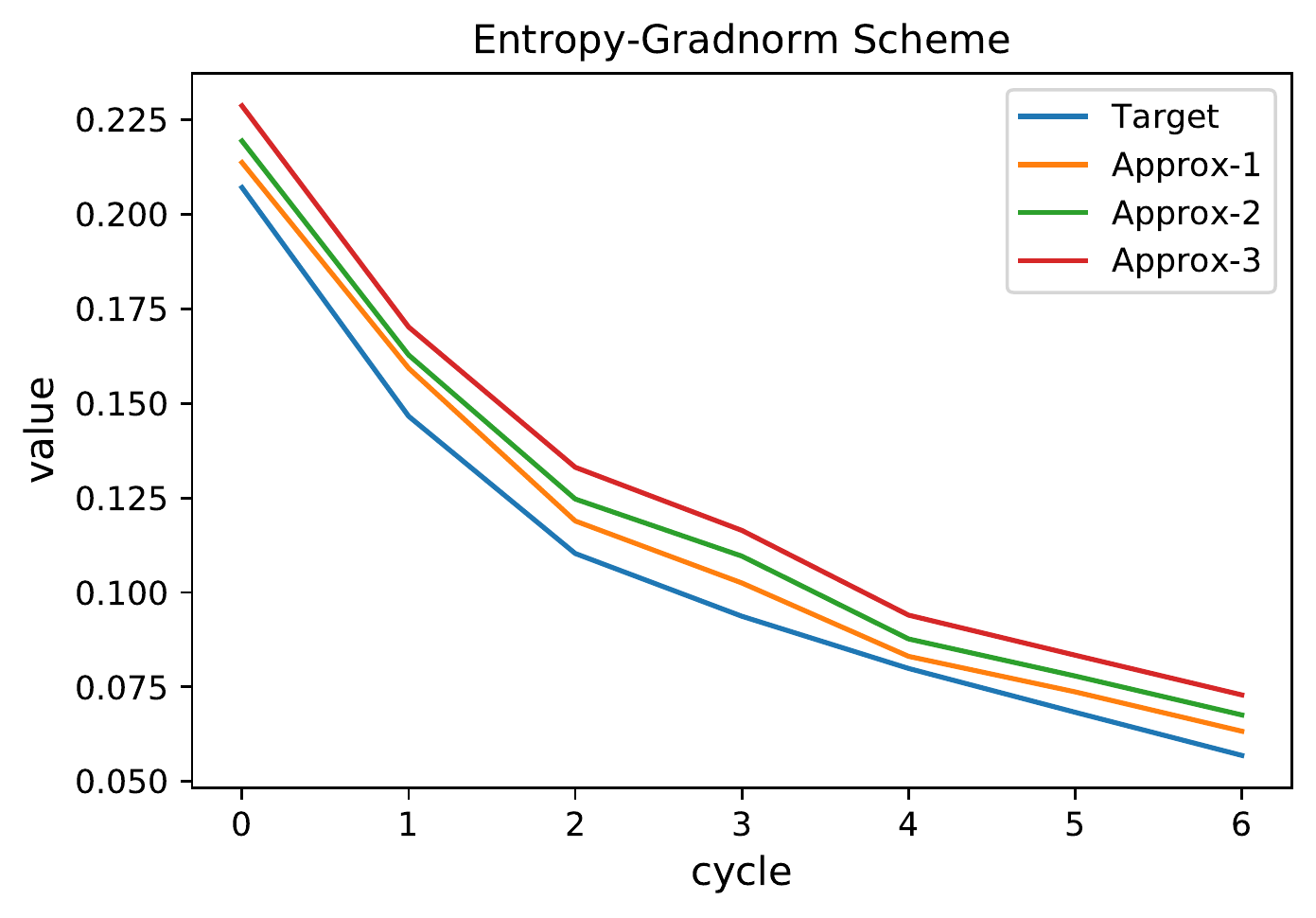}
    \caption{How the upper-bound changes after the three approximations in Eq.~\ref{eq-approx1} and Eq.~\ref{eq-approx2} based on the expected-gradnorm scheme (\textbf{left}) and entropy-gradnorm scheme (\textbf{right}), respectively, evaluated on Cifar10.}
    \label{fig:approx}
\end{figure*}

\subsection{Quantitative evaluation of the Bounds in Eq.~\ref{eq-approx1} and Eq.~\ref{eq-approx2}}
Here we quantitatively validate the bounds in Eq.~\ref{eq-approx1} and Eq.~\ref{eq-approx2}, derived from the original target in Eq.~\ref{eq-target}, 
by presenting two types of evaluations.

Firstly, we evaluate how the samples selected by Eq.~\ref{eq-approx2} are consistent with that selected by Eq.~\ref{eq-target}. Note that Eq.~\ref{eq-approx2} is the data selection criterion used in our algorithm. Specifically, for Eq.~\ref{eq-target} we compute the influence of each $x$ by \textit{temporarily} “observing” its label. At each AL cycle, we acquire the top 2500 samples selected by Eq.~\ref{eq-target} and Eq.~\ref{eq-approx2}, respectively. Then we check the proportion of the overlapped samples. As shown in Table~\ref{table:appro1}, more than 90\% of the selected samples are consistent between the original (Eq.~\ref{eq-target}) and approximated criterion (Eq.~\ref{eq-approx2}). 
Therefore, Eq.~\ref{eq-approx2} is a reasonable surrogate of 
Eq.~\ref{eq-target} in AL context. \textit{Note that only in this experiment, for the purpose of evaluation, we assume test data is available in order to compute Eq.~\ref{eq-target}.}

\begin{table}[!bh]
\centering
\small
\begin{tabular}{|l|ccccccc|}
\hline
\multirow{2}{*}{Eq.} & \multicolumn{7}{c|}{Cycle} \\
\cline{2-8}
 & 0 & 1 & 2 & 3 & 4 & 5 & 6 \\
\hline
\ref{eq-target} & 2500 & 2500 & 2500 & 2500 & 2500 & 2500 & 2500  \\
\ref{eq-approx2} & 2290 & 2377 & 2396 & 2432 & 2449 & 2438 & 2482  \\
\hline
\end{tabular}
\caption{How the samples selected by Eq.~\ref{eq-approx2} are consistent with that selected by Eq.~\ref{eq-target}, evaluated on Cifar10.}
\label{table:appro1}
\end{table}

Secondly, we illustrate how the upper-bound changes after each approximation in Eq.~\ref{eq-approx1} and Eq.~\ref{eq-approx2}. Since we actually use either Eq.~\ref{eq:ExpectedGN} or Eq.~\ref{eq:EntropyGN} as the loss to compute gradient norm for unlabeled data, these upper-bounds are computed based on the proposed Expected-Gradnorm scheme and Entropy-Gradnorm scheme, respectively. Specifically, after each AL cycle, we compute four terms. 
\begin{itemize}
    \item \textbf{Target}: \small $\sum_j \nabla_\theta L(T^{c+1}(x_{j}))^\top H_{\theta}^{-1} \nabla_\theta L(T^{c+1}(x))$ \normalsize in line 2 of Eq.~\ref{eq-target}, which is the exact influence to test loss. 
    \item \textbf{Approx-1}: \small $||\sum_j \nabla_\theta L(T^{c+1}(x_{j}))^\top H_{\theta}^{-1} \nabla_\theta L(T^{c+1}(x))||$ \normalsize in line 2 of Eq.~\ref{eq-approx1}. 
    \item \textbf{Approx-2}: a further approximation in line 4 of Eq.~\ref{eq-approx1}: \small $||\nabla_\theta L(T^{c+1}(x))|| \cdot ||\sum_j H_{\theta}^{-1} \nabla_\theta L(T^{c+1}(x_{j})) ||$ \normalsize .
    \item \textbf{Approx-3}: \small $||\nabla_\theta L(T^{c}(x))|| \cdot ||\sum_j H_{\theta}^{-1} \nabla_\theta L(T^{c+1}(x_{j})) ||$ \normalsize in line 2 of Eq.~\ref{eq-approx2}.
\end{itemize}
Note that when using the Expected-Gradnorm,
all $L$ in above four terms will be the expected loss defined in Eq.~\ref{eq:ExpectedGN}. 
When using the Entropy-Gradnorm, all $L$ in the four terms will be the entropy loss defined in Eq.~\ref{eq:EntropyGN}.
For each term, we use all unlabeled samples to compute an average value. The results in Fig.~\ref{fig:approx} demonstrate that the bounds in the three approximations do not deviate much from the target in Eq.~\ref{eq-target}, indicating the reliability of these derived bounds.

\subsection{Better Generalization}
Here, we compare the generalization of the task model that is trained on the data selected by the different AL methods. Specifically, after each AL cycle, we compute the gap between the training and test accuracy. The smaller the better for this value, since a larger one indicates that the task model suffers more from over-fitting. As shown in Table \ref{table:overfit}, our method yields the smallest gap, demonstrating its superior generalization ability.

\begin{table}[!bh]
\centering
\small
\begin{tabular}{|l|cccccc|}
\hline
\textbf{budget (\%)} & 15 & 20 & 25 & 30 & 35 & 40 \\
\hline

\textbf{mc-dropout} & 16.14 & 10.25 & 9.47 & 8.77 & 8.24 & 7.35  \\

\textbf{core-set} & 14.59 & 12.73 & 10.88 & 9.95 & 8.92 & 8.31  \\

\textbf{vaal} & 15.01 & 12.24 & 11.08 & 9.66 & 8.57 & 8.08  \\

\textbf{ll4al} & 13.83 & 10.7 & 8.89 & 8.78 & 7.64 & 7.23  \\

\textbf{sraal} & 14.09 & 11.55 & 10.24 & 8.62 & 8.01 & 6.94  \\

\textbf{exp-gn} & \textbf{12.74} & 10.04 & 8.06 & 7.05 & \textbf{6.09} & 5.66  \\

\textbf{ent-gn} & 13.05 & \textbf{9.9} & \textbf{7.76} & \textbf{6.79} & 6.29 & \textbf{5.65}  \\

\hline

\end{tabular}
\vspace{0.2em}
\caption{Gap (\%) between training and test accuracy after each AL cycle, evaluated on Cifar10. Annotation budget of 10\% is ignored since all the methods use the same randomly selected initial data for the first cycle training.}
\label{table:overfit}
\vspace{-0.5cm}
\end{table}

\section{Conclusion}
In this paper, we theoretically analyze the connection between data selection and the test performance of the task model used in active learning. We prove that selecting unlabeled data of higher gradient norm can reduce the upper-bound of the test loss. We propose two independent schemes to compute gradient norm and a universal active learning framework to leverage the schemes. We conduct extensive experiments on various benchmark datasets and 
the promising results validate our theoretical findings and the proposed schemes. 

\section*{Acknowledgement}
We thank all the reviewers for their valuable comments that help us to improve this paper. 

\bibliography{aaai22}

\clearpage
\input{appendix}

\end{document}

%% file: appendix.tex
\section{Appendix}

\input{proof_gradient_reduction}

\subsection{A.2 Quantitative Evaluations for Eq. (5)}
For quantitative evaluations, in each AL cycle we investigate how many selected samples have reduced gradient norm in next cycle. We use the proposed \textit{Expected-Gradnorm} and \textit{Entropy-Gradnorm} schemes respectively to compute the gradient norm for each selected sample. We show the results in Table~\ref{table:Eq5_ExpectdGN} and Table~\ref{table:Eq5_EntropyGN}, respectively. As illustrated, more than 90\% (e.g. 2488/2500) of all the selected samples have reduced gradient norm in all the cycles, suggesting that our derived bound is reasonable.

\begin{table}[th]
\centering
\begin{tabular}{|c|c|c|c|c|c|c|}
\hline
\textbf{Cycle} &1&2&3&4&5&6\\
\hline
\textbf{\#} & 2488&2481&2466&2439&2386&2255 \\ 
\hline
\end{tabular}
\caption{Number of the selected samples that satisfy $||\nabla_\theta L(T^{c+1}(x))|| \leq ||\nabla_\theta L(T^{c}(x))||$. Cycle 0 is ignored since right after cycle 0 model $T^1$ is not available yet. All the values are out of 2500, which is the number of the samples selected in each cycle. In this evaluation (on Cifar10), the \textit{Expected-Gradnorm} scheme is used. }
\label{table:Eq5_ExpectdGN}
\end{table}

\begin{table}[th]
\centering
\begin{tabular}{|c|c|c|c|c|c|c|}
\hline
\textbf{Cycle} &1&2&3&4&5&6\\
\hline
\textbf{\#} & 2487&2477&2467&2447&2388&2250 \\ 
\hline
\end{tabular}
\caption{Number of the selected samples that satisfy $||\nabla_\theta L(T^{c+1}(x))|| \leq ||\nabla_\theta L(T^{c}(x))||$. Cycle 0 is ignored since right after cycle 0 model $T^1$ is not available yet. All the values are out of 2500, which is the number of the samples selected in each cycle. In this evaluation (on Cifar10), the \textit{Entropy-Gradnorm} scheme is used. }
\label{table:Eq5_EntropyGN}
\end{table}

\begin{figure*}[th]
    \centering
    \includegraphics[width=0.45\linewidth]{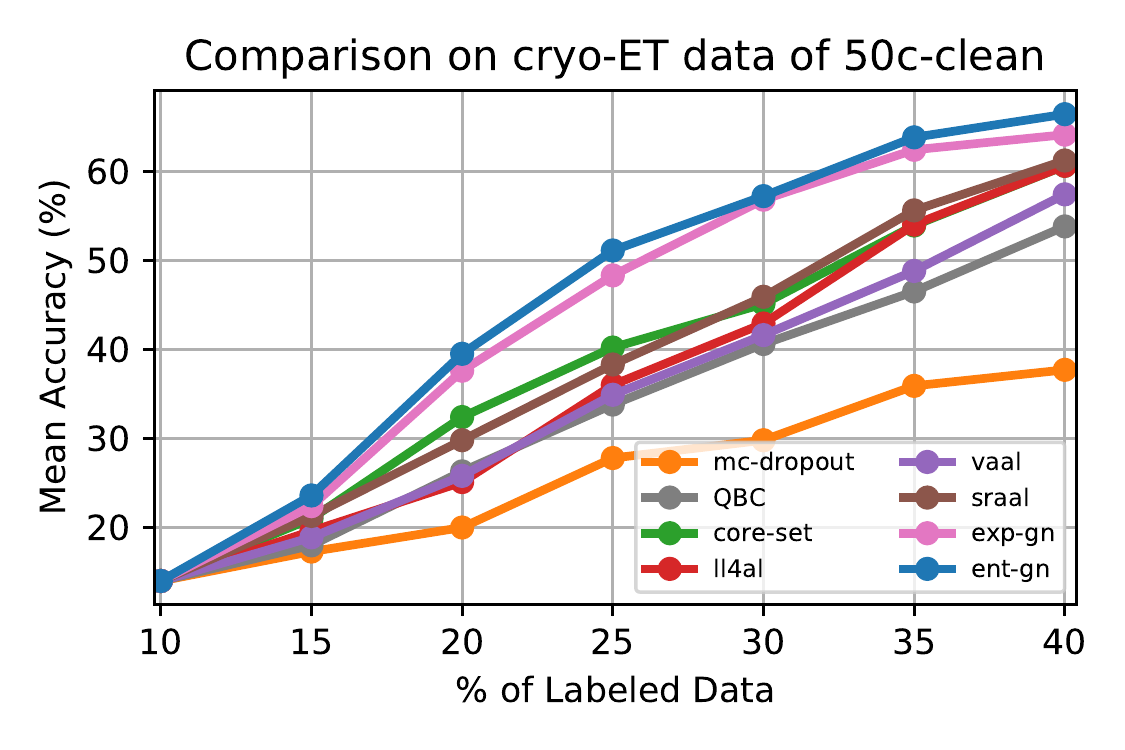}
    \includegraphics[width=0.45\linewidth]{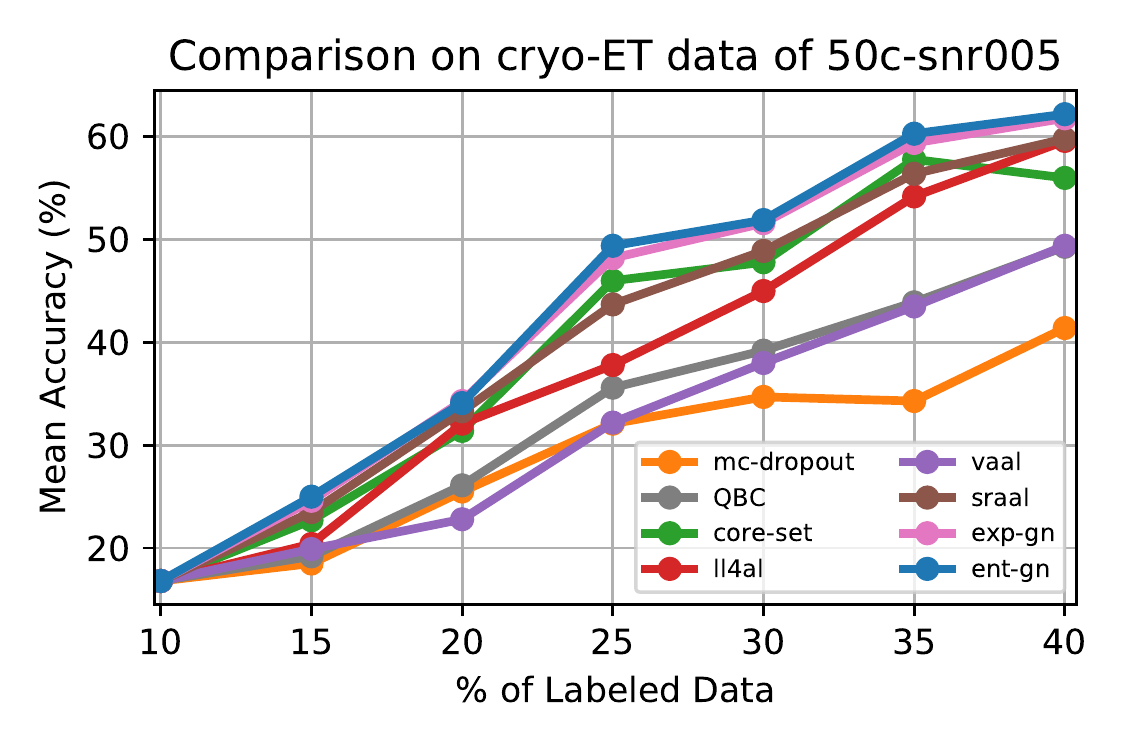}
    
    \caption{Classification performance of the AL methods on cryo-ET data: 50c-clean (\textbf{left}) and 50c-snr005 (\textbf{right}).}

    \label{fig:cryoET}
\vspace{-0.3cm}
\end{figure*}

\subsection{A.3 Benefits of Selecting Unlabeled Data of Higher Gradient Norm}
We present 
an intuitive explanation of why selecting samples of higher gradient norm. 
Without loss of generality, we follow the notations in \textbf{Methodology} of the paper. Assuming that a removed sample will lead to an increased test loss, our goal is to select such a sample $x$ which maximizes $\sum_j I_{loss}(x, x_{j})=\frac{1}{n} \sum_j \nabla_\theta L(f_{\theta}(x_{j}))^\top H_{\theta}^{-1} \nabla_\theta L(f_{\theta}(x))$.
Denote $\frac{1}{n} \nabla_\theta L(f_{\theta}(x_{j}))^\top H_{\theta}^{-1}$ by \bm{${c}_j$}, which is the projected (by $H_{\theta}^{-1}$) gradient of a test sample $x_j$ and approximately independent of an individual training sample $x$. As for $x$, we express its gradient \bm{$g_x$} with the direction \bm{$\vec{g}_x$} and magnitude \bm{$\|g_x\|$} as \bm{$g_x =\|g_x\| \vec{g}_x$}. Then $\sum_j I_{loss}(x, x_{j})$ can be reformulated as
\begin{equation}
\small
\begin{split}
  \sum_j \left<\bm{{c}_j}, \bm{g_x}\right>=\bm{\|g_x\|}\sum_j \left<\bm{{c}_j}, \bm{\vec{g}_x}\right>\enspace.
\end{split}
\end{equation}
While it is obvious that both higher \bm{$\|g_x\|$} and $\sum_j \left<\bm{{c}_j}, \bm{\vec{g}_x}\right>$
will increase the test loss, paying attention to \bm{$\|g_x\|$} can be favorable due to two reasons. (1) As each \bm{${c}_j$} normally has a different direction, it is inefficient to find the optimal \bm{$\vec{g}_x$}, especially in a high dimensional feature space. However, solely increasing \bm{$\|g_x\|$} is easy and has a universal effect over the space of \bm{${c}_j$}. (2) When a greedy algorithm is adopted, which is common in AL methods, selecting data according to the direction may be sub-optimal in terms of the diversity of a sample set. This is because the selected samples are prone to have similar directions (i.e. samples with the same label and similar visual features normally have similar gradient directions). In contrast, increasing \bm{$\|g_x\|$} has an implicit effect of selecting a set of more diverse samples. 


\subsection{A.4 Limitation of the Expected-Gradnorm Scheme}
As discussed in section \textbf{Expected-Gradnorm Scheme} of the paper, this scheme is not an ideal solution for complex tasks in which assuming label is infeasible, such as the semantic segmentation task. 
Here, we further explain this limitation. In this scheme, for each unlabeled sample, we need to successively assume each possible label as the ground-truth label of the sample. For classification problems, there will be $N$ possibilities, where $N$ is the number of classes in a dataset (e.g. $N=100$ in Cifar100 \cite{krizhevsky2009learning}). In practice, we only need to run a loop of $N$ iterations. However, for semantic segmentation problems, if the image resolution is $H\times W$, then there will be $C^{H\times W}$ possibilities when making label assumptions, where $C$ denotes the number of classes for each pixel. For example, in the Cityscapes \cite{cityscapes} task, we have $C=19$, $H=688$, and $W=688$, which means for each unlabeled sample we need to run a loop of $19^{473344}$ iterations to estimate its expected empirical loss, which is infeasible in practice. Therefore, we propose another scheme in section \textbf{Entropy-Gradnorm Scheme} of the paper to compute gradient norm without assuming labels.

\subsection{A.5 Cryo-ET Challenges}
\label{cryo-sec}
AL has been demonstrated significantly useful in domain tasks \cite{kuo2018cost, gorriz2017cost,konyushkova2017learning}, since the annotation cost in such tasks is extremely high. We explore using AL on cryo-ET (cryo-Electron Tomography) subtomogram classification \cite{chen2017convolutional}, which is an important domain task in computational biology. This experiment also demonstrates the robustness of our method to noise since part of the used data contains high level of noise. 

\noindent
\textbf{Datasets.} We evaluate our method on two simulated cryo-ET datasets consisting of subtomograms. The first one, namely \textit{50c-clean}, is noise free. The second one, namely \textit{50c-snr005}, contains noise with a SNR (signal-noise-ratio) of 0.05. Each dataset includes 24000 training and 1000 test samples, uniformly distributed across 50 classes. One can refer to \cite{gubins2019shrec} for the details of the data generation. 

\noindent
\textbf{Model Selection.} Unlike RGB pixel image, each subtomogram is composed of voxels (e.g. $32\times32\times32$). Therefore, we customize ResNet-18 with 3D operations (e.g. 3D convolutions) as the task model. It is employed for all the compared methods. The VAE helper in vaal \cite{sinha2019variational} and sraal \cite{zhang2020state}, and the loss prediction module in ll4al \cite{yoo2019learning} need to be re-designed to fit the cryo-ET data.
Therefore, extra effort is indispensable if using these methods. 

\noindent
\textbf{Implementation Details.}
We train the task model for 100 epochs with an initial learning rate of 0.1, decayed by a factor of 0.1 at the $80^{th}$ epoch. The SGD optimizer \cite{bottou2010large} is adopted. The momentum and weight decay rate are set to 0.9 and $5\times10^{-4}$, respectively. 

\noindent
\textbf{Results and Analysis.}
As illustrated in Fig. \ref{fig:cryoET}, our method yields consistently better results than the baselines, demonstrating its reliability. When there is high level of noise in the data (i.e. SNR of 0.05), our method still yields superior performance, demonstrating its robustness to noise.

\begin{table*}[th]
\centering
\begin{tabular}{|l|c|c|c|c|c|c|c|}
\hline
\textbf{Method} & mc-dropout & core-set & vaal & ll4al & sraal & expected-gradnorm & entropy-gradnorm\\
\hline
\textbf{ResNet-50} & 91.23 & 90.12 & 90.97 & 92.32 & 92.68 & \textbf{94.1} & \textbf{93.16} \\ 
\hline

\end{tabular}
\vspace{0.5em}
\caption{Classification accuracy (\%) of ResNet-50 that is trained on the data selected by ResNet-18. 20,000 samples in the training set of Cifar10 are selected by each AL method, respectively. 
}
\label{table:verydeep}
\end{table*}

\begin{table*}[th]
\centering
\begin{tabular}{|l|c|c|c|c|c|c|c|c|c|c|}
\hline
\textbf{Class (\# of samples)} & 0 (1744)& 1 (5099) & 2 (4149)& 3 (2882)& 4 (2523) \\

\hline
\textbf{Accuracy} & 96.85 & 97.51 & 97.06 & 93.82 & 97.23   \\
\hline
\textbf{Class (\# of samples)} & 5 (2384)& 6 (1977)& 7 (2019)& 8 (1660)& 9 (1595) \\
\hline 

\textbf{Accuracy} & 95.22 & 96.81 & 95.84 & 96.39 & 95.17   \\
\hline

\end{tabular}
\vspace{0.5em}
\caption{Class-wise classification accuracy (\%) of our \textbf{expected-gradnorm} scheme on the test data of SVHN.}
\label{table:class-wise1}
\end{table*}

\begin{table*}[!th]
\centering
\begin{tabular}{|l|c|c|c|c|c|c|c|c|c|c|}
\hline
\textbf{Class (\# of samples)} & 0 (1744)& 1 (5099) & 2 (4149)& 3 (2882)& 4 (2523)  \\

\hline
\textbf{Accuracy} & 97.19 & 96.96 & 97.42 & 93.34 & 97.42   \\
\hline
\textbf{Class (\# of samples)} & 5 (2384)& 6 (1977)& 7 (2019)& 8 (1660)& 9 (1595) \\
\hline

\textbf{Accuracy} & 95.85 & 96.46 & 96.04 & 96.02 & 94.55   \\
\hline

\end{tabular}
\vspace{0.5em}
\caption{Class-wise classification accuracy (\%) of our \textbf{entropy-gradnorm} scheme on the test data of SVHN.}
\label{table:class-wise2}
\end{table*}

\begin{table*}[!th]
\centering
\begin{tabular}{|l|c|c|c|c|c|c|}
\hline
\textbf{Method} & mc-dropout & core-set & vaal  & ll4al & expected-gradnorm & entropy-gradnorm \\
\hline
\textbf{Cifar10} & 4.38 & 11.13 & 15.21 & 2.51 & 2.33 & 2.29 \\ 
\hline
\textbf{Cifar100} & 4.49 & 11.25 & 15.83 & 2.18 & 2.42 & 2.36   \\
\hline
\textbf{SVHN} & 6.35 & 26.61 & 22.76 & 3.42 & 3.28 & 3.17  \\
\hline
\textbf{Caltech101} & 1.43 & 1.01 & 17.58 & 0.86 & 0.93 & 0.9  \\
\hline

\end{tabular}
\vspace{0.5em}
\caption{Training time (in hours) of the AL methods on the four classification datasets. 
All the results are measured on a Nvidia RTX 2080Ti GPU with an Intel(R) Xeon(R) Gold 5117 CPU.}
\label{table:efficiency}
\end{table*}

\subsection{A.6 Ablation Study}
\label{ablation}

\noindent
\textbf{Training Deeper Model with Selected Data.}
Here, we study whether the data selected by ResNet-18 \cite{he2016deep} in AL context can effectively train a deeper model from scratch, such as ResNet-50 \cite{he2016deep}. As shown in Table \ref{table:verydeep}, ResNet-50 achieves the best test performance when it is trained on the data selected by our method, further demonstrating that our selected data can greatly benefit model training.

\noindent
\textbf{Class-wise Performance for Imbalanced Data.}
Here, we investigate whether the promising performance of our method is dominated by the majority classes in an imbalanced dataset, such as SVHN \cite{svhn}.
We run our method for 7 AL cycles and evaluate the final trained model on the test set of SVHN.
As shown in Table \ref{table:class-wise1} and \ref{table:class-wise2}, our method works well on both majority (e.g. class 1 and 2) and
minority classes (e.g. class 8 and 9). In Table \ref{table:class-wise2}, the highest accuracy of 97.42\% is achieved in both class 2 and 4, whereas the latter includes much fewer samples than the former. These observations demonstrate the robustness of our method to imbalanced datasets.

\subsection{A.7 Time Efficiency}
We compare the training time
of our method with that of the baselines in Table \ref{table:efficiency}. 
\textit{Note that in this work, we count the data selection time as part of the training time.}
On all the datasets, we run each AL method for 7 cycles (i.e. annotation budget varies from 10\% to 40\% with an incremental size of 5\%). The implementation details remain the same as described in section \textbf{A.8} of this Appendix. 
We observe that our method is on par with ll4al \cite{yoo2019learning}, but much more efficient than the others. 
This is because in our method only the task model needs to be trained, whereas in vaal \cite{sinha2019variational} two auxiliary models (i.e. a VAE \cite{kingma2013auto} and a discriminator) need to be trained in addition to the task model. 
In core-set \cite{sener2018active}, one needs to solve the K-center problem iteratively when selecting unlabeled data, thus time consuming, while in our method one only needs a feed-forward and a back-propagate step for each unlabeled sample. The time expense of mc-dropout \cite{gal2016dropout} is also higher than ours since it needs a large number of feed-forward steps (e.g. 1000 times) when estimating uncertainty for each unlabeled sample. 
Note that vaal is much slower than the others on Caltech101 \cite{caltech}. This is because Caltech101 consists of higher resolution images (e.g. $300\times200$), and hence a smaller batch size of 16 is appropriate for vaal to avoid CUDA memory issues, whereas the others can use 64 as the batchsize. The memory issue of vaal is due to the training of the multiple auxiliary models on the bigger images in Caltech101.

\subsection{A.8 Training Details}
\noindent
Instead of selecting data from an entire unlabeled pool \{$X_U$\}, we follow the practice in \cite{beluch2018power} to construct a random subset \{$R_U$\}$\subset$\{$X_U$\}, and select data from this subset. Assuming we select $K$ samples in each AL cycle, generally we have $len(\{R_U\})\approx 10\times K$, where $len(\cdot)$ computes the length of a set. We apply this rule to all the experiments in this work.

\noindent
\textbf{Image Classification.} Here we present the training specifications of the image classification experiments in this paper. For the two Cifar and SVHN datasets, we train the task model for 200 epochs with an initial learning rate of 0.1 and a batch size of 128. The learning rate is decayed by a factor of 0.1 at the $160^{th}$ epoch. For Caltech101, we train the task model for 50 epochs and the initial learning rate is set to 0.01, which is decayed by 0.1 at the $40^{th}$ epoch. The batch size is set to 64. We use the SGD optimizer \cite{bottou2010large} with a momentum and weight decay rate of 0.9 and $5\times10^{-4}$, respectively. For ImageNet, we train the task model for 100 epochs with a batch size of 64. We use the Adam optimizer \cite{kingma2014adam} with a learning rate of 0.1 through the entire training process.

\noindent
\textbf{Semantic Segmentation.} Here we specify the training settings of the semantic segmentation experiments in this paper. We train the task model for 50 epochs with a batch size of 8. The Adam optimizer \cite{kingma2014adam} is used with a learning rate of $1\times10^{-4}$, which will not be decayed during training.

\subsection{A.9 More Discussions of Related Work}
\noindent
The authors in \cite{settles2007multiple} designed the query by considering model change, \textit{nevertheless, only considering model change is not a reliable solution for AL, since incorporating an outlier can also result in a big model change but it will not benefit test performance.}
Moreover, they did not explain how model change would impact test performance. 
On the contrary, in our work we consider a direct impact on \textbf{test loss} (Eq. (5)). By selecting training data of higher gradient norm, we show that the upper-bound of test loss can be lowered. In addition, their scheme is not suitable for complex tasks in which assuming label is impractical (e.g. segmentation). Therefore, we propose another scheme to compute gradient norm with entropy (\textbf{Entropy-Gradnorm}). 
Note that \textit{in our work entropy is not used to sample data}. Instead, \textit{we use entropy as a type of loss to compute gradient norm}, needed in Eq. (5) which establishes a \textbf{\textit{direct}} connection between selected training data and test loss in AL. Therefore, \textit{our main contribution lies in interpreting how selected training data in AL \textbf{directly} influences test performance (in form of loss). To our best knowledge, it has never been explored for deep AL by previous literature, due to the need of model re-training (intractable for deep neural networks (DNNs)).}

Our work differs essentially from \cite{roy2001toward} in the following aspects. (i) \emph{Their method has not been extended to DNNs,} while we are the first to develop a scheme to select the most influential training data with DNNs based on the influence function \cite{koh2017understanding} which formulates the impact of a training point on the prediction of a test point for DNNs. (ii) Their method needs considering all possible labels, making it unsuitable for problems in which assuming label is impractical (e.g. segmentation). \textit{We design the \textbf{Entropy-Gradnorm} scheme that can be easily used for this scenario.} (iii) Their method needs re-training the model for multiple times for data selection. Although several strategies can be used to speed up the process, it is still compute-intensive when deployed for DNNs. Instead, \textit{we only train the model once,} and take the feed-forward operation during data selection, making our method more efficient.

At the time of the acceptance of this paper, we notice that there is a concurrent work \cite{liu2021influence} related to our research. We exploit first-order derivative as the criterion for data selection, whereas they use both first-order and second-order derivative (i.e. in Hessian). In their method, it is not necessary to explicitly compute the inverse of the Hessian, instead the authors take an approximation based on implicit Hessian-vector products (HVPs), as suggested in \cite{koh2017understanding}. 
\textit{The main difference between the two methods is: we use gradient magnitude (norm) for data selection, whereas they use both magnitude and direction.} We think \textit{only using magnitude may have an advantage in terms of data diversity.} We refer readers to the item (2) in section \textbf{A.3} of this Appendix for the discussion.

%% file: proof_gradient_reduction.tex
\subsection{A.1 Proof of the Average Gradient Norm Reduction}
Here we theoretically explain
the rationale of bounding 
$\|\nabla_{\theta}L(T^{c+1}(x))\|$ with $\|\nabla_{\theta}L(T^{c}(x))\|$ in Eq.(5) of the paper. 
Specifically, we provide a theoretical support that gradient norm of average loss (or expected loss over data distribution) is guaranteed to reduce under reasonable assumptions for loss function. 

\begin{proof}
We firstly define the loss function as $f(w)$ and assume $f$ is twice continuously differentiable w.r.t. the parameters $w$. For each iteration $t$, we compute the updated parameters as $w\prime = w - \eta \nabla f(w)$ by performing gradient descent. Denoting $u=w_{\tau}=w+\tau(w\prime-w)=w-\tau\eta\nabla f(w)$, we have 

\begin{equation}
\begin{aligned}
\nabla f(w\prime) &= \nabla f(w-\eta\nabla f(w)) \\
&=\nabla f(w)+\int_{w \to w\prime}\nabla^2 f(u)du \\
&=\nabla f(w)-\eta(\int_{0}^{1}\nabla^2 f(w_{\tau})d\tau)\nabla f(w) \\
&= \nabla f(w)-\eta H \nabla f(w).
\end{aligned}
\end{equation}

The following analysis relies on a widely acknowledged assumption: a deep neural network is locally convex and smooth, which has been discovered by both theoretical and empirical studies~\cite{rister2017piecewise,arora2018stronger,li2018visualizing}. Assuming that $f$ is locally $\mu$-strongly convex and $L$-smooth at each $w$, we have $\mu I \preceq H = \nabla^2 f(w) \preceq LI$. Then, the gradient norm is guaranteed to reduce if we use a small learning rate $\eta$ satisfying $\eta \le 2\mu/L^2$. The proof is as follows,
\begin{equation}
\begin{aligned}
\|\nabla f(w\prime)\|^2 &= \|\nabla f(w)\|^2+\eta(\eta\|H\nabla f(w)\|^2 \\ & \ \ - 2 \nabla f(w)^T H \nabla f(w)) \\
&\le \|\nabla f(w)\|^2+\eta(2\frac{\mu}{L^2}\|L^2 \nabla f(w)\|^2 \\ & \ \ - 2 \nabla f(w)^T \mu I \nabla f(w)) \\
&= \|\nabla f(w)\|^2.
\end{aligned}
\end{equation}

\end{proof}